\documentclass{article}

\usepackage[nonatbib, final]{nips_2016}

\usepackage[utf8]{inputenc} % allow utf-8 input
\usepackage[T1]{fontenc}    % use 8-bit T1 fonts
\usepackage{hyperref}       % hyperlinks
\usepackage{url}            % simple URL typesetting
\usepackage{booktabs}       % professional-quality tables
\usepackage{amsfonts}       % blackboard math symbols
\usepackage{nicefrac}       % compact symbols for 1/2, etc.
\usepackage{microtype}      % microtypography

\usepackage{tikz}
\usepackage{pgfplots}
\usepackage{balance}
\usetikzlibrary{shapes,arrows}

\title{Adversarial Evaluation of Dialogue Models}
\author{
  Anjuli Kannan \\
  Google Brain \\
  \texttt{anjuli@google.com} \\
  \And
  Oriol Vinyals \\
  Google DeepMind \\
  \texttt{vinyals@google.com} \\ 
}

\begin{document}

\maketitle

\begin{abstract}

%% Old abstract - focused on eval:
%% Substantial progress has been made in fully data-driven dialogue systems
%% through the application of RNN encoder-decoder models; however, evaluation
%% remains a significant challenge.  We propose that one measure of a dialogue
%% model's quality is how easily its samples are confused with human-generated
%% samples.  We implement this metric by training an RNN to discriminate the
%% dialogue model's samples from the training distribution.  Testing this metric
%% on a production-scale dialogue system, we find that the discriminating RNN is
%% able to distinguish the two distributions over 60\% of the time, and exploits
%% some of the same major weaknesses that humans have observed in same system.
%% In future work we propose to incorporate this signal into training loss.

The recent application of RNN encoder-decoder models has resulted in
substantial progress in fully data-driven dialogue systems, but evaluation
remains a challenge. 
An adversarial loss could be a way to directly evaluate the extent to which
generated dialogue responses sound like they came from a human.  This could
reduce the need for human evaluation, while more directly evaluating on
a generative task.  In this work, we investigate this idea by training an RNN to
discriminate a dialogue model's samples from human-generated samples.  Although
we find some evidence this setup could be viable, we also note that many issues
remain in its practical application.  We discuss both aspects
and conclude that future work is warranted.

\end{abstract}

\section{Introduction}

Building machines capable of conversing naturally with humans is an open
problem in language understanding. Recurrent neural networks (RNNs) have
drawn particular interest for this problem, typically in the form of an
encoder-decoder architecture.  One network ingests an incoming message (a
Tweet, a chat message, etc.), and a second network generates an outgoing
response, conditional on the first network's final hidden state.  This sort of
approach has been shown to improve significantly over both a statistical machine translation baseline \cite{sordoni} and traditional rule-based chatbots \cite{vinyals_conversation}.

However, evaluating dialogue models remains a significant challenge.  While
perplexity is a good measure of how well a model fits some data, it does not
measure performance at a particular task.  N-gram-based measures such as BLEU,
while useful in translation, are a poor fit for dialogue models because two
replies may have no n-gram overlap but equally good responses to a given
message.  Human evaluation may be ideal, but does not scale well, and can also
be problematic in applications
like Smart Reply \cite{kannan}, where data cannot be viewed by humans.

This work investigates the use of an adversarial evaluation method for dialogue
models.  Inspired by the success
of generative adversarial networks (GANs) for image generation
(\cite{goodfellow}, and others), we propose that one measure of a model's quality is how easily its output is distinguished from a human's output.   As an initial exploration, we
take a fully trained production-scale conversation model deployed as part of
the Smart Reply system (the "generator"), and, keeping it fixed, we
train a second RNN (the "discriminator") on the following task: given an incoming
message and a response, it must predict whether the response was sampled from
the generator or a human.  Our goal here is to understand whether an adversarial
setup is viable either for evaluation.

We find that a discriminator can in fact distinguish the model output from
human output over 60\% of the time.  Furthermore, it seems to uncover the
major weaknesses that humans have observed in the system: an incorrect
length distribution and a reliance on familiar, simplistic replies such as
"Thank you".  Still,
significant problems with the practical application of this method remain.
We lack evidence that a model with lower discriminator accuracy (i.e., that fools it) necessarily
would be better in human evaluation as well.

We present here the details of our analysis, as well as further discussion of
both merits and drawbacks of an adversarial setup.  We conclude that additional
investigation is warranted, and lay out several suggestions for that work.

\subsection{Related work}

Much recent work has employed RNN encoder-decoder models to translate from
utterance to response (\cite{sordoni}, \cite{serban}, \cite{vinyals_conversation}).
Work in \cite{li2} has used policy gradient
but the rewards are manually defined as useful conversational properties such
as non-redundancy.  Evaluation remains a significant challenge
\cite{liu}.

The adversarial setup we describe is inspired by work on GANs for image
generation \cite{goodfellow}; however, we apply the concept to dialogue
modeling, which raises the challenges of sequential inputs/outputs and
conditional generation.  To support our aim of understanding the discriminator,
we also do not train the generator and discriminator jointly.

An adversarial loss for language understanding
is also used in \cite{vilnis} as a means of evaluation;  however, the metric
is not applied to any real world task, nor are the properties of the
discriminator itself explored and evaluated, as we will do in this work.

\section{Model}

Like a GAN, our architecture consists of a
generator and a discriminator; however, these are two
separate models which are not trained to a single objective.

The generator is a sequence-to-sequence model, consisting of an RNN encoder
and an RNN decoder.  Given a corpus of message
pairs $(\mathbf{o}, \mathbf{r})$ where $\mathbf{o}$, the original message,
consists of tokens $\{o_1, ..., o_n\}$ and $\mathbf{r}$, the response
message, consists of tokens
$\{r_1, ..., r_m\}$, this model is trained to maximize the total log probability
of observed response messages, given their respective original messages:

\[ \sum_{(\mathbf{o}, \mathbf{r})}  \log P(r_1, ..., r_m | o_1, ..., o_n)\]

The discriminator is also an RNN, but has only an encoder followed by a binary
classifier.  Given a corpus of message pairs and scores $(\mathbf{o}, \mathbf{r}, y)$ , where
$y = 1$ if $\mathbf{r}$ was sampled from the training data and $0$ otherwise, this
model is trained to maximize:

\[ \sum_{(\mathbf{o}, \mathbf{r}, y)}  \log P(y| o_1, ..., o_n, r_1, ..., r_m )\]

%% TODO: describe variant where makes a pred at each token?

\section{Experiments}

\subsection{Data and training}

We investigate the proposed adversarial loss using a corpus of email reply
pairs $(\mathbf{o}, \mathbf{r})$.  The generator is trained on the same data
and in the same manner as the production-scale model that is deployed
as part of the Smart Reply feature in  Inbox by Gmail \cite{kannan}.
\footnote{In particular, from \cite{kannan}: "All email data (raw data, preprocessed
data and training data) was encrypted. Engineers could only
inspect aggregated statistics on anonymized sentences that
occurred across many users and did not identify any user."}

The discriminator is then trained on a held out set of the email corpus.
For half the pairs $(\mathbf{o}, \mathbf{r})$ in the held out set, we leave
the example unchanged and assign a 1.  For the other half we replace
$\mathbf{r}$ with a message $\mathbf{r'}$ that has been
sampled from the generator, and assign the pair $(\mathbf{o}, \mathbf{r'})$
a score of $0$. Then the discriminator is trained as described in the previous
section.

\subsection{Discriminator performance}

We observe that the discriminator can distinguish between generator
samples and human samples, conditional on an original message, 62.5\% of the
time.  This in itself may be somewhat unexepcted: one may expect that since the
discriminator is only as powerful as the generator, it would not be able to
distinguish its distribution from the training distribution.  A full
precision-recall curve is shown in Figure ~\ref{fig:length}.

\subsection{Comparison with perplexity}

\begin{figure}
\centering
\resizebox{0.5\columnwidth}{0.25\columnwidth}{%
\begin{tikzpicture}
\begin{axis}[
    enlargelimits=false,
]
\addplot+[
    only marks,
    mark size=0.7pt]
table[]
{length.dat};
\end{axis}
\end{tikzpicture}
\begin{tikzpicture}
\begin{axis}[
    xmin=0, xmax=1.0,
    ymin=0, ymax=1.0,
]
\addplot+[
    only marks,
    mark size=0.7pt]
table[]
{pr.dat};
\end{axis}
\end{tikzpicture}
}%
\caption{\textbf{a. (left)} Discriminator score vs length.  \textbf{b. (right)} Recall vs precision}
\label{fig:length}
\end{figure}
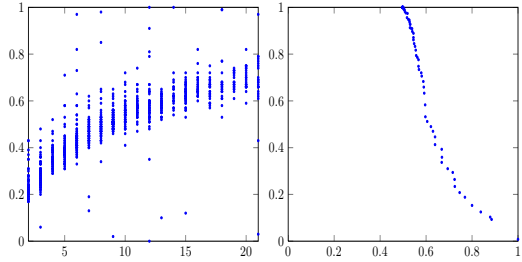

\begin{table}
\begin{tabular}{|p{3.0cm}|p{1.0 cm} |p{4.5 cm} |p{4.5 cm} |}
\hline
Original & Length & Top responses (discriminator) & Top responses (generator) \\
\hline
Actually , yes , let 's & 3 & Oh alright . & Ok thank you\\
move it to Monday . & & That 's fine & That 's fine \\
      & & Good ... . & thank you !\\
\hline
Are you going to  & 1 & Ya & Yes \\
Matt 's party on & & Maybe & No \\
Sunday ?      & & Yeah & yes \\
\hline
\end{tabular}
\caption{Comparison of ranking of responses of the same length by discriminator
score and generator score.  The discriminator favors less common
language like "Ya" over "Yes" and consistently omits "Thank you" from its top responses.}
\label{fig:samples}
\end{table}

A qualitative analysis shows that the discriminator objective favors different
features than the generator objective.  To demonstrate this, we sample 100
responses from the generator for each of 100 donated email messages.  These are
then ranked according to both the discriminator score and the generator's
assigned log likelihood, and the two rankings are compared.

First, we see that the discriminator's preferences are strongly correlated with
length (Figure ~\ref{fig:length}).  This is relevant because it has been
previously documented that sequence-to-sequence models have a length bias
\cite{sountsov}.  The discriminator relies too heavily on this signal, favoring
longer responses even when they are not internally coherent.
Still, it is noteworthy that it identifies something humans have
documented as a key weakness of the model \cite{vinyals_conversation}.

The discriminator does not assign equal probability to all responses of the
same length.  When comparing responses of the same length, we find that
it has a significantly different ranking than the likelihood assigned by the
generator, with an average Spearman's correlation of -0.02.
Broadly speaking we find that the discriminator has less preference for the
most common responses produced by the generator, things like "Thank you!" and
"Yes!" (Table ~\ref{fig:samples}).  The
lack of diverse generated language has  been documented as
a weakness of these dialogue models in \cite{kannan} and \cite{li}, both of which
incorporate significant post-processing and re-ranking to overcome this noted
weakness.  As with length, the discriminator's preference for rarer language
does not necessarily mean it is favoring better responses; it is noteworthy only
in that it shows signs of detecting the known weakness of the generator.
Future work might incorporate minibatch discrimination \cite{salimans} to more
explicitly address the diversity weakness.

\section{Discussion}
\label{sec:discussion}

In this research note we investigated whether the discriminator in GANs can be
employed for automatic evaluation of dialogue systems. We see a natural
progression towards using discriminators:

\begin{enumerate}
  \item{Ask humans to evaluate each single system published in a consistent
        manner.  Though ideal, it would also be time consuming and
        prohibitively expensive.}
  \item{Annotate a large dataset of dialogues, learn a “critic” (e.g., a neural
        network), and use it to score any new system (so that extra human
        labour would not be required). However, this critic would likely not
        perform well when evaluated off-policy, and overfitting could occur,
        as researchers may naturally find its weaknesses.}
  \item{Use the discriminator of a GAN as a proxy for giving
        feedback akin to a human. Since training such a critic would
        be simple for any dialogue system, each research group could provide
        theirs and any new system could be evaluated with a variety of
        discriminators.}
\end{enumerate}

The last item is the simplest, and it is what we have explored in this work. Our
preliminary work suggests that the critic we
trained on a production-quality dialogue system is able to automatically
find some of the previously identified weaknesses when
employing probabilistic models -- sequence length and
diversity. It also succeeds in identifying real vs generated responses of a
highly tuned system.

However, as with GANs, more needs to be understood and using discriminators
alone won’t solve the evaluation challenges of dialogue systems. Despite the
fact that GANs do not use any extra information than what’s already present
in the training dataset, some have argued that it is a better loss than likelihood
\cite{yu}.  Still, there remains a tension between what we train the
discriminator on (samples) and what we typically use in practice (the maximally
likely response, or some approximation of it).  Discriminators have a harder
time when sampling versus using beam search, but this conflicts with human
observations that some amount of
search typically is useful to get the highest quality responses.  Further work
is required to understand if and how discriminators can be applied in this domain.

% \section{Conclusion}

\small
\bibliographystyle{abbrv}
\bibliography{main}

\begin{thebibliography}{10}

\bibitem{vilnis}
S.~R. Bowman, L.~Vilnis, O.~Vinyals, A.~M. Dai, R.~Jozefowicz, and S.~Bengio.
\newblock Generating sentences from a continuous space.
\newblock In {\em arXiv preprint arXiv:1511.06349}, 2015.

\bibitem{goodfellow}
I.~Goodfellow, J.~Pouget-Abadie, M.~Mirza, B.~Xu, D.~Warde-Farley, S.~Ozair,
  A.~Courville, and Y.~Bengio.
\newblock Generative adversarial nets.
\newblock In {\em Proceedings of NIPS}, 2014.

\bibitem{kannan}
A.~Kannan, K.~Kurach, S.~Ravi, T.~Kaufmann, B.~Miklos, G.~Corrado, and et~al.
\newblock Smart reply: Automated response suggestion for email.
\newblock In {\em Proceedings of KDD}, 2016.

\bibitem{li}
J.~Li, M.~Galley, C.~Brockett, J.~Gao, and B.~Dolan.
\newblock A diversity-promoting objective function for neural conversation
  models.
\newblock In {\em Proceedings of NAACL-HLT}, 2016.

\bibitem{li2}
J.~Li, W.~Monroe, A.~Ritter, M.~Galley, J.~Gao, and D.~Jurafsky.
\newblock Deep reinforcement learning for dialogue generation.
\newblock In {\em arXiv preprint arXiv:1606.01541}, 2016.

\bibitem{liu}
C.-W. Liu, R.~Lowe, I.~V. Serban, M.~Noseworthy, L.~Charlin, and J.~Pineau.
\newblock How not to evaluate your dialogue system: An empirical study of
  unsupervised evaluation metrics for dialogue response generation.
\newblock In {\em EMNLP}, 2016.

\bibitem{salimans}
T.~Salimans, I.~Goodfellow, W.~Zaremba, V.~Cheung, A.~Radford, and Z.~Chen.
\newblock Improved techniques for training gans.
\newblock In {\em arXiv preprint arXiv:1606.03498}, 2016.

\bibitem{serban}
I.~V. Serban, A.~Sordoni, Y.~Bengio, A.~Courville, and J.~Pineau.
\newblock Hierarchical neural network generative models for movie dialogues.
\newblock In {\em arXiv preprint arXiv:1507.04808}, 2015.

\bibitem{sordoni}
A.~Sordoni, M.~Galley, M.~Auli, C.~Brockett, Y.~Ji, M.~Mitchell, J.-Y. Nie,
  J.~Gao, and B.~Dolan.
\newblock A neural network approach to context-sensitive generation of
  conversation responses.
\newblock In {\em Proceedings of NAACL-HLT}, 2015.

\bibitem{sountsov}
P.~Sountsov and S.~Sarawagi.
\newblock Length bias in encoder decoder models and a case for global
  conditioning.
\newblock In {\em Proceedings of EMNLP}, 2016.

\bibitem{vinyals_conversation}
O.~Vinyals and Q.~V. Le.
\newblock A neural conversation model.
\newblock In {\em ICML Deep Learning Workshop}, 2015.

\bibitem{yu}
L.~Yu, W.~Zhang, J.~Wang, and Y.~Yu.
\newblock Seqgan: Sequence generative adversarial nets with policy gradient.
\newblock In {\em arXiv preprint arXiv:1609.05473}, 2016.

\end{thebibliography}

\end{document}